\documentclass{scspaperproc}

\usepackage{latexsym}
\usepackage{graphicx}
\usepackage{mathptmx}

\usepackage{amsmath}
\usepackage{amsfonts}
\usepackage{amssymb}
\usepackage{amsbsy}
\usepackage{amsthm}

\usepackage[pdftex,colorlinks=true,urlcolor=blue,citecolor=black,anchorcolor=black,linkcolor=black,bookmarks=false]{hyperref}

\usepackage{hyphenat}
\newtheoremstyle{scsthe}
{8pt}
{8pt}
{\it}
{}
{\bf}
{.}
{.5em}
{}

\theoremstyle{scsthe}

\begin{document}

%
%

\pagestyle{fancyplain}

\thispagestyle{plain}
\firstPageHead{}

\chead{\fancyplain{}{\itshape\small Yu and Haskins \vspace{8pt}}}

\rhead{}
\cfoot{}
\renewcommand{\headrulewidth}{0pt} 

\makeatletter
\let\@internalcite\cite
\def\cite{\def\@citeseppen{-1000}%
    \def\@cite##1##2{(##1\if@tempswa , ##2\fi)}%
    \def\citeauthoryear##1##2##3{##1 ##3}\@internalcite}
\def\citeNP{\def\@citeseppen{-1000}%
    \def\@cite##1##2{##1\if@tempswa , ##2\fi}%
    \def\citeauthoryear##1##2##3{##1 ##3}\@internalcite}
\def\citeN{\def\@citeseppen{-1000}%
    \def\@cite##1##2{##1\if@tempswa, ##2)\else{}\fi}%
    \def\citeauthoryear##1##2##3{##1 (##3)}\@citedata}
\def\citeA{\def\@citeseppen{-1000}%
    \def\@cite##1##2{(##1\if@tempswa , ##2\fi)}%
    \def\citeauthoryear##1##2##3{##1}\@internalcite}
\def\citeANP{\def\@citeseppen{-1000}%
    \def\@cite##1##2{##1\if@tempswa , ##2\fi}%
    \def\citeauthoryear##1##2##3{##1}\@internalcite}
\def\shortcite{\def\@citeseppen{-1000}%
    \def\@cite##1##2{(##1\if@tempswa , ##2\fi)}%
    \def\citeauthoryear##1##2##3{##2 ##3}\@internalcite}
\def\shortciteNP{\def\@citeseppen{-1000}%
    \def\@cite##1##2{##1\if@tempswa , ##2\fi}%
    \def\citeauthoryear##1##2##3{##2 ##3}\@internalcite}
\def\shortciteN{\def\@citeseppen{-1000}%
    \def\@cite##1##2{##1\if@tempswa, ##2\else{}\fi}%
    \def\citeauthoryear##1##2##3{##2 (##3)}\@citedata}
\def\shortciteA{\def\@citeseppen{-1000}%
    \def\@cite##1##2{(##1\if@tempswa , ##2\fi)}%
    \def\citeauthoryear##1##2##3{##2}\@internalcite}
\def\shortciteANP{\def\@citeseppen{-1000}%
    \def\@cite##1##2{##1\if@tempswa , ##2\fi}%
    \def\citeauthoryear##1##2##3{##2}\@internalcite}
\def\citeyear{\def\@citeseppen{-1000}%
    \def\@cite##1##2{(##1\if@tempswa , ##2\fi)}%
    \def\citeauthoryear##1##2##3{##3}\@citedata}
\def\citeyearNP{\def\@citeseppen{-1000}%
    \def\@cite##1##2{##1\if@tempswa , ##2\fi}%
    \def\citeauthoryear##1##2##3{##3}\@citedata}
%
%
%
\def\@citedata{%
    \@ifnextchar [{\@tempswatrue\@citedatax}%
                  {\@tempswafalse\@citedatax[]}%
}

\def\@citedatax[#1]#2{%
\if@filesw\immediate\write\@auxout{\string\citation{#2}}\fi%
  \def\@citea{}\@cite{\@for\@citeb:=#2\do%
    {\@citea\def\@citea{, }\@ifundefined
       {b@\@citeb}{{\bf ?}%
       \@warning{Citation `\@citeb' on page \thepage \space undefined}}%
{\csname b@\@citeb\endcsname}}}{#1}}%

%
\def\@citex[#1]#2{%
\if@filesw\immediate\write\@auxout{\string\citation{#2}}\fi%
  \def\@citea{}\@cite{\@for\@citeb:=#2\do%
    {\@citea\def\@citea{, }\@ifundefined
       {b@\@citeb}{{\bf ?}%
       \@warning{Citation `\@citeb' on page \thepage \space undefined}}%
{\csname b@\@citeb\endcsname}}}{#1}}%

%
\def\@biblabel#1{}
\makeatother

\newdimen\bibindent
\bibindent=.25in

\def\thebibliography#1{\section*{\refname}\list
   {}{\settowidth\labelwidth{[#1]}
   \leftmargin \bibindent
   \itemindent -\bibindent
   \listparindent \itemindent
	 \itemsep 4pt
   \parsep 0pt
   \usecounter{enumi}}
   \def\newblock{}
   \sloppy
   \sfcode`\.=1000\relax}

\setlength{\baselineskip}{12.7pt}

\def\SCSconferenceacro{}

\def\SCSpublicationyear{}

\def\SCSconferencedates{}

\def\SCSconferencevenue{}

\title{KNN, An Underestimated Model for Regional Rainfall Forecasting}

\author{
\\
Ning Yu \\ [12pt]
Department of Computing Sciences \\
SUNY Brockport \\
350 New Campus Drive \\
Brockport, NY, USA 14420 \\
nyu@brockport.edu \\
\and
 \\
Timothy Haskins \\ [12pt]
Department of Computing Sciences \\
SUNY Brockport \\
350 New Campus Drive \\
Brockport, NY, USA 14420\\
thask1@brockport.edu
}

\maketitle

\section*{Abstract}

Regional rainfall forecasting is an important issue in hydrology and meteorology. This paper aims to design an integrated tool by applying various machine learning algorithms, especially the state-of-the-art deep learning algorithms including Deep Neural Network, Wide Neural Network, Deep and Wide Neural Network, Reservoir Computing, Long Short Term Memory, Support Vector Machine, K-Nearest Neighbor for forecasting regional precipitations over different catchments in Upstate New York. Through the experimental results and the comparison among machine learning models including classification and regression, we find that KNN is an outstanding model over other models to handle the uncertainty in the precipitation data. The data normalization methods such as ZScore and MinMax are also evaluated and discussed.

\textbf{Keywords:} rainfall forecasting, k-nearest neighbor, deep and wide neural network, reservoir computing, long short term memory. 

\section{Introduction}
\label{sec:intro}

New York historically had sufficient precipitation until recently, with intense drought occurring over the 2016 growing season, especially in western New York \cite{todaro_2018}. The observed precipitation in 2016 was less than normal, with shortfalls of 4-8 inches being common in the 90 days leading up to the drought watch. Accurate rainfall forecasting is important for planning in agriculture and other relevant activities.  \par

Although a number of modern algorithms and applications have been used to forecast rainfall, there are two categories of approaches to solve the problem. The first category models the underlying physical principles for the rainfall process. However, it is thought not feasible limited by the complex climatic system in various spatial and temporal dimensions. A second category is based on the data mining and pattern recognition, which attempts to mine rainfall patterns and learn the knowledge from numerous features and a large volume of data. Historical meteorological data including precipitation data are used to feed and train the recognition model and further predict the evolution of other storms. In the recent years, with the advance of big data and deep learning technology, this type of approach has drawn more attentions from researchers. It does not require a thorough understanding of the climatic laws but it does need the data modeling for data mining and pattern recognition \cite{LUK_etal_2001}. \par

There are a number of reported research that have used machine learning algorithms to solve problems in hydrology. These modern machine learning algorithms include Deep Neural Network (DNN), Wide Neural Network (WNN), Deep and Wide Neural Network (DWNN), Reservoir Computing (RC), Long Short Term Memory (LSTM), Support Vector Machine (SVM), K-Nearest Neighbor (KNN), and so forth. \par

As a type of Artificial Neural Networks (ANN), DNN can emulate the process of the human nervous system and have proven to be very powerful in dealing with complicated problems, such as computer vision and pattern recognition. Moreover, DNN is computationally robust even when input data contains lots of uncertainty such as errors and incomplete, as such examples are very common in rainfall prediction.  \par

A pioneer two-dimensional ANN model \cite{French_etal_1992} has been considered as the first Neural Network (NN) application to simulate complex geophysical processes and forecast the rainfall, although it was limited by many aspects including insufficient neural network configurations and a mathematical rainfall simulation model that was used to generate the input data. More applications of NN model were developed later. \cite{Koizumi_1999} utilized an ANN model for radar, satellite and weather-station data together to provide better performance than the persistence forecast, the linear regression forecasts, and the numerical model precipitation prediction. By comparing several short-time rainfall prediction models including linear stochastic auto-regressive moving-average (ARMA) models, artificial neural networks (ANN) and the non-parametric nearest-neighbours method, \cite{toth_2000} revealed that those ANN models provide a significant improvement in real-time flood forecasting. \cite{LUK_etal_2001} developed three types of Artificial Neural Networks including time delay neural networks for rainfall forecasting. They revealed that the rainfall time series have very short term memory characteristics. \par 


\cite{hung_2009} applied ANN model for real time rainfall forecasting and flood management in Bangkok, Thailand. The developed ANN model was fed with meteorological parameters (relative humidity, air pressure, wet bulb temperature and cloudiness), the rainfall at the point of forecasting and rainfall at the surrounding stations that fueled the ANN model to predict the rainfall at any moment. \par

In recent years, the performance of three Artificial Neural Networks (ANN) approaches were evaluated in the forecasting of the monthly rainfall anomalies for Southwestern Colombia \cite{canchala_2020}. A forecasting performance was found suitable for the regional rainfall forecasting in Southwestern Colombia and the combination of ANN approaches had demonstrated the ability to provide useful prediction information for the decision-making. \par

Google researchers have invented Wide \& Deep Neural Network (WDNN) model for recommender systems originally, where it contains two components: wide neural network and deep neural network. The wide neural network (WNN) was designed to effectively memorize the feature interactions through a wide set of cross-product feature transformations. The deep neural networks (DNN) were used to generalize better to unseen feature combinations through low-dimensional dense embeddings learned for the sparse features \cite{cheng2016wide}.  WDNN has been used in various applications, such as clinical regression prediction for Alzheimer patients \cite{polsterl2019wide}. In the latest publication, \cite{bajpai2021deep} analyzed and evaluated three deep learning approaches, one dimensional Convolutional Neutral Network, Multi-layer Perceptron and Wide Deep Neural Networks for the prediction of summer monsoon rainfall in Indian state of Rajasthan and found that Deep and Wide Neural Network (DWNN) can achieve the best performance over the rest two deep learning methods. \par

As a type of recurrent neural network (RNN), Long Short-Term Memory neural network (LSTM) has been studied by many researchers in recent literature. \cite{liu2020applicability} used LSTM to simulate rainfall–runoff relationships for catchments with different climate conditions. The LSTM model then was coupled with the k-nearest neighbor (KNN) algorithm as an updating method to further boost the performance of LSTM method. LSTM-KNN model was validated by comparing the single LSTM and the recurrent neural network (RNN). \cite{chhetri2020deep} studied LSTM, Bidirectional LSTM (BLSTM), and Gated Recurrent Unit (GRU), another variant of RNN. LSTM recorded the best Mean Square Error (MSE) score outperformed other six different existing models including Linear Regression, Multi-Layer Perceptron (MLP), Convolutional Neural Network (CNN), GRU, and BLSTM. They further modified the model by combining LSTM and GRU to carry out monthly rainfall prediction over a region in the capital of Bhutan. \par

As one of machine learning models, K-nearest neighbor model has shown a promising performance in climate prediction. \cite{jan2008seasonal} applied KNN for climate prediction by using the historical weather data of a region such as rain, wind speed, dew point, temperature, etc. \cite{huang2017novel} developed a K-nearest neighbor (KNN) algorithm to offer robustness the irregular class distribution of the precipitation dataset and made a sound performance in precipitation forecast. However, they did not show the comparison with other advanced machine learning algorithms such as deep learning, limiting its application on the rainfall forecasting. \cite{yang2020geca} developed an Ensemble–KNN forecasting method based on historical samples to avoid uncertainties caused by modeling inaccuracies. \cite{hu2013emd} proposed a model that combined empirical mode decomposition (EMD) and the K-nearest neighbor model to improve the performance of forecasting annual average rainfall. \cite{wu2009novel} proposed a KNN nonparametric regression model used to improve an ANN-based model. \par

Like multilayer perceptrons and radial basis function networks, support vector machines can be used for pattern classification and nonlinear regression. SVM is found to be a significant technique to solve many classifications problem in the last couple of years. \cite{hasan2015support} exhibited a robust rainfall prediction technique using Support Vector Regression (SVR). Support Vector Machine based approaches also have illustrated the quite capability for rainfall forecasting. Support vector regression combined with Singular Spectrum Analysis (SSA) has shown its efficiency for monthly rainfall forecasting in two reservoir watersheds \cite{bojang2020linking}. \par

The Reservoir Computing framework has been known for over a decade as a state-of-the-art paradigm for the design of RNN \cite{coble2020reservoir}. Among the models of RC instances, the echo state network (ESN) represents a type of the most widely known schemes with a strong theoretical support and various applications \cite{gallicchio2017deep}. \cite{yen2019application} used the ESN and the DeepESN algorithms to analyze the meteorological hourly data in Taiwan and showed that DeepESN was better than that by using the ESN and other neuronal network algorithms. Backpropagation Algorithms and Reservoir Computing in Recurrent Neural Networks were adopted to predict the complex spatiotemporal dynamics \cite{vlachas2020backpropagation}.  \par

The precipitation prediction models in this study are based on not only the implementation of the state-of-the-art machine learning algorithms but also the design of an integrated tool for these machine learning algorithms to provide better regional rainfall forecasting.  A variety of machine learning algorithms are applied and compared including DNN, WNN, DWNN, RC, LSTM, SVM, KNN for forecasting regional precipitations over different catchments in Upstate New York. These algorithms were applied to forecast rainfall at multiple locations simultaneously for the next hour based on previously observed precipitation data. The important issues of data normalization and comparison of algorithms are discussed also. \par

\section{Models and Algorithms}
\label{sec:models}

A set of algorithms of DNN, WNN, DWNN, RC, LSTM, SVM, KNN are used to forecast regional rainfall. Both classification and regression models of these algorithms are implemented in our integrated tool. For the classification models, the threshold of rainfall and non-rainfall is 0.01 inch. Any of below the threshold is defined as very light precipitation according to the categories in hydrology. For the regression models, the root of mean-squared error (RMSE) and mean-squared error (MSE) are used to measure the performance of these regression models. We describe these main models and algorithms used in our study as following subsections.  \par

\subsection{Wide Neural Network}
\label{subsec:wnn}

The core of a wide neural network is a generalized linear model as shown in \eqref{eq:linear}. 
\begin{equation} \label{eq:linear}
y=w^{T}x+b, 
\end{equation}
where $x$ is a vector of $d$ features, $w$ is the weights or parameters, and $b$ is the bias. The $d$  feature set may be raw input features or transformed features. A cross-product transformation is utilized to capture the interactions between the binary features, which adds a non-linear factor to the generalized linear model, as shown in \eqref{eq:transform}.
\begin{equation} \label{eq:transform}
\phi_{k}(x)=\prod_{i=1}^{d}x_{i}^{c_{ki}}, c_{ki}\in \{0,1\}, 
\end{equation}
where $c_{ki}$ is 1 if the $i$-th feature is part of the $k$-th transformation $\phi_{k}$ and 0 if the $i$-th feature is not part of the $k$-th transformation $\phi_{k}$. The generalized linear models with cross-product feature transformations can memorize these exception rules with much fewer parameters \cite{cheng2016wide}.\par
 
\subsection{Deep Neural Network}
\label{subsec:dnn}
\begin{figure}[htb]
{
\centering
\includegraphics[width=0.50\textwidth]{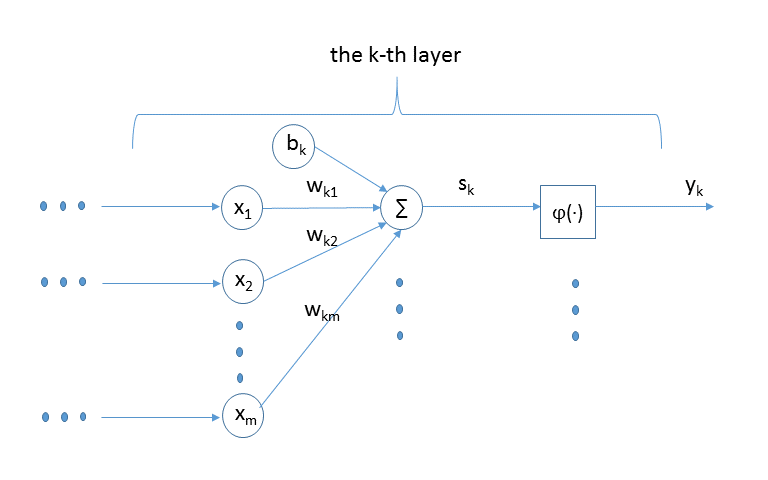}
\caption{Structure of perceptron}\label{fig:perceptron}
}
\end{figure}

Artificial neural Networks are generalized models of biological neuron system. Deep neural network is a type of artificial neural networks, which contains a number of layers and can massively distribute processing over layers of neurons or perceptrons. Each neuron or perceptron is the elementary unit containing a simple linear summing function and an activation function as shown in \autoref{fig:perceptron}. The summing function is illustration in \autoref{eq:summing}:
\begin{equation} \label{eq:summing}
s_{k}=\sum_{j=1}^{m}w_{kj}x_{j}+b_{k}, 
\end{equation}
where $k$ denotes the identifier of the $k$-th layer, $x_j$ is the input of node $j$, and $b_k$ is the bias of layer $k$. The sigmoid function is often used as an activation function as shown in \autoref{eq:sigmoid}:
\begin{equation} \label{eq:sigmoid}
y_k(s)=\varphi (s)=\frac{1}{1+e^{-s}}. 
\end{equation}

\subsection{Wide and Deep Neural Network}
\label{subsec:wdnn}

\begin{figure}[htb]
{
\centering
\includegraphics[width=0.50\textwidth]{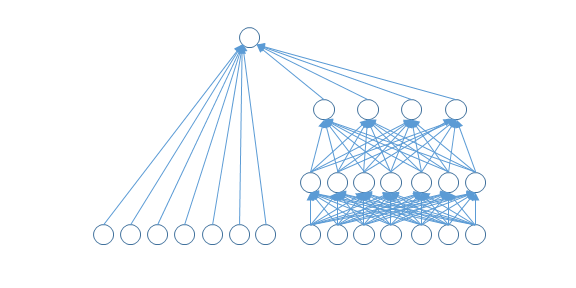}
\caption{An illustration of DWNN}\label{fig:dwnn}
}
\end{figure}

The Wide and Deep Neural Network is composed of a wide neural network and a deep neural network using a weighted sum of their output log odds as the prediction and a common logistic loss function for a joint training. In the joint training, the wide component uses a small number of cross-product feature transformations to complements the weaknesses of the deep neural network \cite{cheng2016wide}. \par

The deep component is a feed-forward neural network, as shown on the right side of \autoref{fig:dwnn} while the wide component is a linear model based neural network, as shown on the left side of \autoref{fig:dwnn}. The low-dimensional vectors are fed into the hidden layers of a neural network in the forward pass according to \autoref{eq:hidden}:
\begin{equation} \label{eq:hidden}
y_{k+1} = \varphi(w_{k} y_{k} + b_{k}),
\end{equation}
where $k$ is the layer number and $\varphi$ is the activation function, $y_k$, $w_k$, and $b_k$ are the output, model weights, and bias at the $k$-th layer. \par

\subsection{K-Nearest Neighbors}
KNN algorithm is a non-parametric method that can approximate its outcome by averaging the observations in the same neighbourhood. The neighbourhood is determined by the distance between the observations and the input point. The distance can be measured by many methods such as Manhattan distance, Euclidean distance, and Minkowski distance. Euclidean distance is calculated as the square root of the sum of the squared differences between a new point (x) and an existing point (y) to calculate the distance between the new point and each training point. \autoref{eq:euclidean} shows the distance measurement:
\begin{equation} \label{eq:euclidean}
D(x,y)=\sqrt{\sum_{i=1}^{n}(x_i-y_i)^2},
\end{equation} 
where $i$ is the $i$-th feature of nodes $x$ and $y$. KNN can be used for both classification and regression. In kNN regression, the output value is the average of the values of k nearest neighbors. In kNN classification, the output value is the most common class among k nearest neighbors. The value of k needs to be set manually or can be chosen or using cross-validation to select the size that minimizes the mean-squared error or maximize the accuracy. \par

\subsection{Reservoir Computing}

\begin{figure}[htb]
{
\centering
\includegraphics[width=0.50\textwidth]{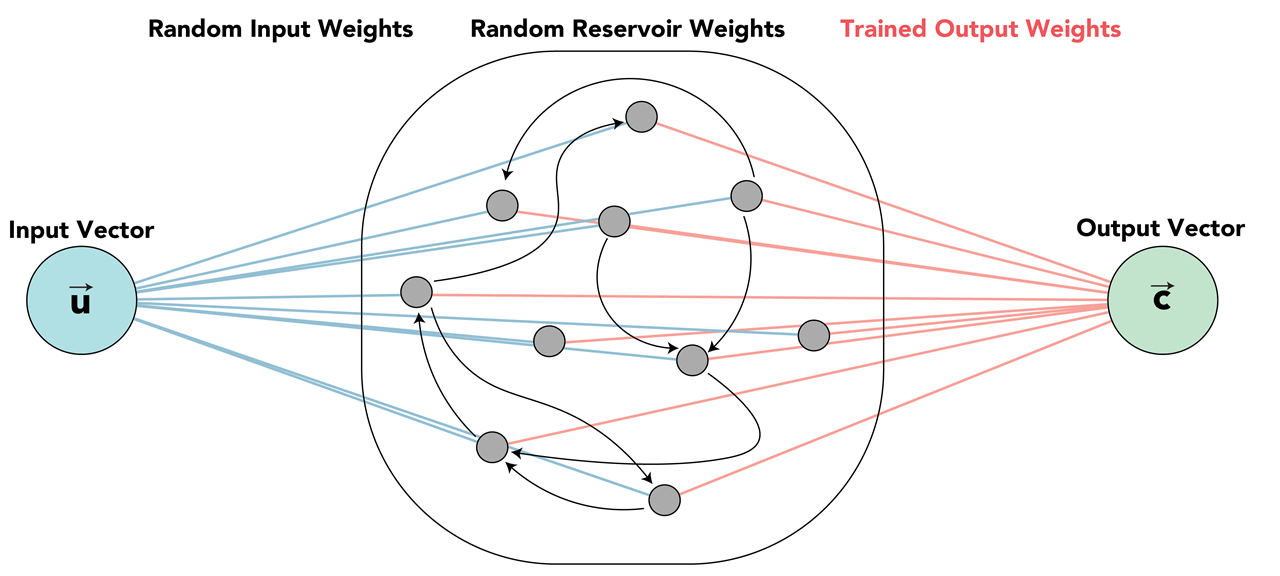}
\caption{An illustration of reservoir computing}\label{fig:reservoir}
}
\end{figure}

The reservoir computing (RC) framework consists of three layers: an input layer, reservoir layer, and output layer as depicted in \autoref{fig:reservoir}. 

The input layer consists of the input weights to each reservoir perceptron. It is a randomly generated matrix of size ($\#$ of properties$)\times (\#$ of perceptrons), with values uniformly distributed between 0 and $ \delta $, where $\delta$ is a scaling factor. This scaling factor allows us to adjust the reservoir computing method to training data of various sizes. The input layer matrix is denoted as $W_{in}$. \par

The reservoir layer is where the actual RNN is held. The current, $i$-th, state of the reservoir is described as a length ($\#$ of perceptrons) vector, referenced as $\vec{r}(i)$. The perceptron weights are described by a connectivity/adjacency matrix, $A$. This matrix is an Erdös-Rényi randomly generated matrix. The eigenvalues of the matrix are normalized, and scaled by a factor, $\rho $, known as the spectral radius. The spectral radius of the connectivity matrix determines how the reservoir adapts to changing dynamics. Usually, a spectral radius of around 0.9 is used in dynamics-prediction applications. The activation function for the reservoir is given in \autoref{eq:activation}. 
\begin{equation}
\label{eq:activation}
\vec{r}(i)=\tanh [\vec{r}(i-1) \times A+\vec{u}(i) \times W{_{in}} ],
\end{equation}
where $\vec{r}(i-1)\times A$ represents the reservoir feedback and is what allows the reservoir to adapt to new input vectors. The $\vec{u}(i) \times W_{in}$ term represents the input to the reservoir. The input term and activation function used varies greatly between different implementations of reservoir computing \cite{coble2020reservoir}.\par
The output layer is how a prediction is obtained from the reservoir. The output layer is a ($\#$ of perceptrons) $ \times $( $\#$ of classes) matrix, referred to as $W_{out}$. The output is obtained by \autoref{eq:output}. 	
\begin{equation}
\label{eq:output}
output=\vec{r}(i)\times W{_{out}} 
\end{equation}
\par

\subsection{Support Vector Machine}
No matter it is for classification or regression, support vector machine (SVM) or support vector regressor (SVR) aims to find out the hyperplane. The difference between SVM and SVR is that one is to segregate the nodes for classification while another is to find the decision boundary along with hyperplane and have the least error rate to fit the model. The hyperplane in a SVM is shown in \autoref{eq:hyperplane}.  \par
\begin{equation}
\label{eq:hyperplane}
y=\vec{w}\cdot x+b
\end{equation}

The learning of the hyperplane is done by training the samples using some kernel. \autoref{eq:svm} shows the prediction of a new input using a kernel. 
\begin{equation}
\label{eq:svm}
f(x)=\sum_{i} a_i K(x_i,x)+b_0,
\end{equation}
where $K$ is the kernel function, $x$ is the input, $x_i$ the support vector, the coefficients $b_0$ and $a_i$ are estimated from the training data through the learning algorithm. A common kernel function $K$ can be linear such as dot product, or non-linear such as Radial basis function kernel (rbf) as shown in \autoref{eq:rbf}, 

\begin{equation}
\label{eq:rbf}
K(x_i,x)=e^{-\frac{\left \| x-x_i \right \|^{2}}{2\delta ^{2}}}, 
\end{equation}
where $\delta$ is a free parameter, and $\left \| x-x_i \right \|^{2}$ is the Euclidean distance.

\subsection{Long Short Term Memory}

\begin{figure}[htb]
{
\centering
\includegraphics[width=0.50\textwidth]{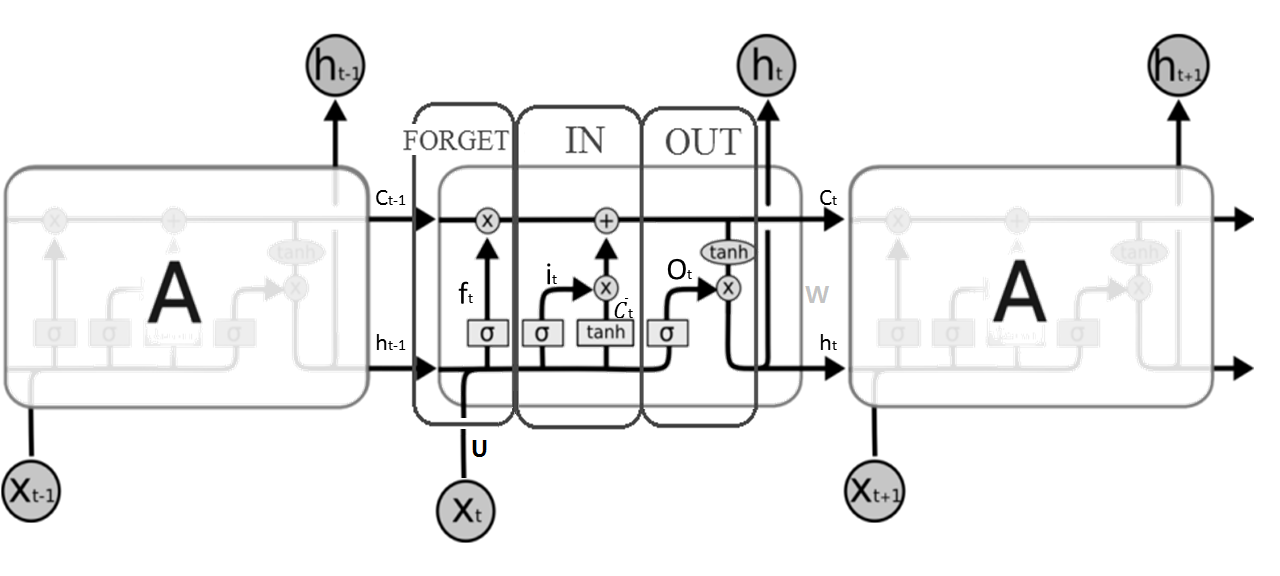}
\caption{An illustration of LSTM structure}\label{fig:lstm}
}
\end{figure}

The LSTM network accepts the output of the previous moment, the current system state, and the current system input, updates the system status through the input gate, the forget gate, and the output gate, then finally outputs the result \cite{hochreiter1997long}. As shown in \autoref{fig:lstm}, the forget gate is $f(t)$ , the input gate is $i(t)$, the output gate is $o(t)$, and the cell state is $C_t$. \par

The forget gate decides the information forgotten and controls the transfer of information from the previous time as shown in \autoref{eq:lstm-ft}: 
\begin{equation}
\label{eq:lstm-ft}
f_t=\delta (U_f x_t+W_f h_{t-1}).
\end{equation}
The input gate determines the current time system input, which acts in front of the forget gate to supplement the latest memory of the network, as shown in \autoref{eq:lstm-it}:
\begin{equation}
\label{eq:lstm-it}
i_t=\delta (U_i x_t+Wi h_{t-1}).
\end{equation}

The candidate layer  $\overline{C}$ is shown in \autoref{eq:lstm-ctp}: 
\begin{equation}
\label{eq:lstm-ctp}
\overline{C_t}=tanh(U_c x_t+W_ch_{t-1}).
\end{equation}

The current cell memory $C_t$ is defined in \autoref{eq:lstm-ct}:
\begin{equation}
\label{eq:lstm-ct}
C_t=f_tC_{t-1}+i_t\overline{C_t}.
\end{equation}

The output gate $O_t$ is shown in \autoref{eq:lstm-ot}:
\begin{equation}
\label{eq:lstm-ot}
O_t=\delta (U_ox_t+W_oh_{t-1}).
\end{equation}

The current cell output $h_t$ is shown in \autoref{eq:lstm-ht}:
\begin{equation}
\label{eq:lstm-ht}
h_t=tanh(C_t)O_t.
\end{equation}


Bidirectional LSTM is viewed as a variant of LSTM. Using bidirectional can receive the input from two directions: one is from past to future and one is from future to past that is different from unidirectional LSTM. Gated Recurrent Unit (GRU) is similar to a LSTM model but it lacks a output gate so that it has fewer parameters than LSTM. It has demonstrated better performance in smaller dataset \cite{cho2014learning}. \par  

\section{Data Preparation}
\label{sec:data}

The data was obtained from Iowa State University’s Iowa Environmental Mesonet which has a portal to download Automated Surface Observing System (ASOS) data from weather stations around the US. Four regional data sets are collected. Rochester data came from the ROC weather station. Buffalo data came from the BUF and IAG weather stations. Syracuse data came from the SYR and FZY weather stations. Albany data came from the ALB and SCH weather stations. With this tool, we were able to collect 11 years of historical hourly weather data from the locations we needed for our project. The features of the data that we collected for each location were; Hourly precipitation in inches, Temperature in Fahrenheit, Dew Point in Fahrenheit, Relative Humidity as a percent, Wind Direction in degrees from North, Wind Speed in knots, Pressure altimeter in inches, Sea Level Pressure in millibar, and Visibility in miles. The target was the Precipitation in inches for the next hour in Rochester. \par

\begin{figure}[htb]
{
\centering
\includegraphics[width=1.0\textwidth]{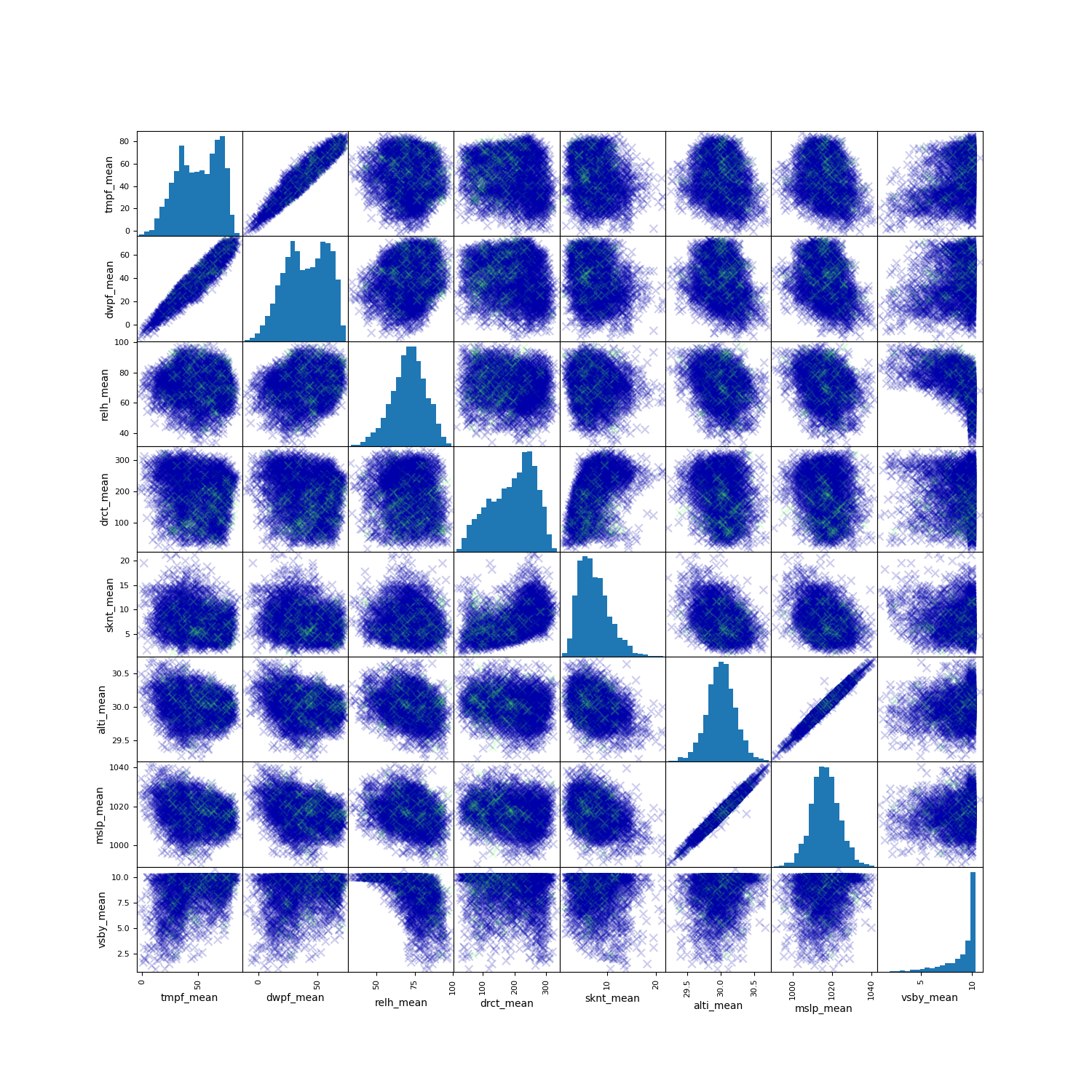}
\caption{The feature matrix in ROC dataset}\label{fig:puzzle}
}
\end{figure}

The plot matrix shown in \autoref{fig:puzzle} shows the relationships between the numeric features for Rochester. There are 2 sets of features with clear linear correlations, that being Temperature and Dew Point as well as Pressure and Pressure at Sea Level. Temperature and Dew point are commonly used in calculating Heat Index so it makes sense that they are strongly related. The two different measurements of air pressure seem nearly identical but as you can see in the chart, there are outliers showing that they are not exactly the same. It is for this reason that we decided to keep both of them. Besides the two sets of strongly correlated features, the other features present a more complicated relationship. In this plot matrix, the data points in which the following day had precipitation above 0 inches are marked in green while the rest of the data points are blue. \par

The raw data was grouped into hour intervals starting with 0 on January 1, 2010, at midnight. These groups were averaged into a single row for all of the features except p01i which is specifically the accumulated precipitation within an hour so the max was taken for each group for the p01i feature.
All missing p01i values were set to 0. The decision was made after looking at random hour intervals whose p01i value was missing and cross-referencing that time frame with Weather.com. In all instances, Weather.com had 0 inches of accumulated precipitation where the ASOS p01i values were missing. For the rest of the features, missing values were interpolated from their previous and future hour interval values. \par
From this cleaning and grouping, two separate datasets were created. One dataset had only data from Rochester. The second dataset had data from Rochester as well as corresponding data from Buffalo, Syracuse, and Albany. The second dataset was created by performing a left join of the Rochester data with a union of the rest of the data. Intervals where Rochester had data but the others did not, had their values interpolated by the same method that was used to previously remove missing values. \par

At this point in the data’s life, two separate versions of the two datasets were made. One for classification where an interval’s target was a binary 0/1 of whether the next interval’s p01i value exceeded 0.01 inches. Another for the regressors where an interval’s target was the next interval’s p01i value. The threshold of 0.01 inches was chosen because although hourly rainfall in inches is already a very small amount, eliminating the smallest of the values reduced noise. \par
For each of the 4 datasets, a min-max normalization and a z-score normalization version were created. This resulted in 6 datasets for classification and 6 datasets for regressors. For all of the classifiers and regressors, the 11 years of hourly data was split into 70\% for training and 30\% for testing. For all except LSTM, the split was made by shuffling the data by a given random state and split after shuffling. In the cases where random states were used to shuffle the data, the random states considered were None, 0, and 42. \par

\section{Experimental Results}
\label{sec:results}
\subsection{Classification Models}
\subsubsection{KNN}
Each of the 6 datasets was crossed with the 3 different random seeds for a total of 18 forms of data. Each of those 18 forms was iterated through a SKLearn KNeighborClassifier 42 times to find the best n\_neighbor value. After 756 iterations the best accuracy was 96.34\% which was obtained from the Rochester, Buffalo, Syracuse, and Albany dataset, normalized by z-score, a random state of None, and 11 n\_neighbors.
\subsubsection{RCC}
Reservoir Computing Classifier was investigated because, in a previous investigation of daily values, RCC had the highest accuracy. For RCC, each of the 6 datasets was crossed with the same 3 random states as before and crossed again with 6 different sizes of the reservoir; 50, 100, 200, 400, 600, 1000. The result of these crosses was 108 unique data forms to train with. The best accuracy was 95.88\% which was obtained from the Rochester only dataset, normalized by z-score, a random state of None, and a reservoir size of 1000.
\subsubsection{SVM}
There were 72 unique Support Vector Machine Classifiers by training against the cross of 6 premade datasets, 4 Kernels, and the same 3 Random States as before. The best accuracy for SVM, and ultimately the best accuracy for all classifiers, was 96.39\% obtained from the ROC+BUF+SYR+ALB, normalized with ZScore, using None as Random State and the rbf kernel.
\subsubsection{DNN}
72 DNN Classifiers were tested and used by training against the cross of the 6 premade datasets, 4 different amounts of hidden layers, and the 3 Random States used from before. The 4 different hidden layers were: 2 layers, 3 layers, 4 layers, and 5 layers. Each of the hidden layers had the number of inputs equal to the number of features which for the ROC datasets was 9 and for the ROC+BUF+SYR+ALB there were 36 inputs at each hidden layer. The best accuracy for DNN was 94.88\% obtained from the ROC+BUF+SYR+ALB, normalized with ZScore, using None as Random State and with 5 hidden layers.

\subsubsection{WNN}
There were 18 unique WNN Classifiers by training against the cross of the 6 premade datasets and the 3 Random States used from before. The best accuracy for WNN was 96.06\% obtained from the ROC+BUF+SYR+ALB, normalized with ZScore, using None as Random State.

\subsubsection{DWNN}
There were 72 unique DWNN Classifiers by training against the cross of the 6 premade datasets, 4 different amounts of hidden layers for the deep aspect, and the 3 Random States used from before. The 4 different hidden layers were; 2 layers, 3 layers, 4 layers, and 5 layers. Each of the hidden layers had the number of inputs equal to the number of features which for the ROC datasets was 9 and for the ROC+BUF+SYR+ALB there were 36 inputs at each hidden layer. The best accuracy for DWNN was 95.91\% obtained from the ROC+BUF+SYR+ALB, normalized with ZScore, using None as Random State and with 4 hidden layers.

\subsubsection{LSTM}
There were 12 unique LSTM Classifiers by training against the cross of the 6 premade datasets, and 2 different sequence lengths. Random states were not used for LSTM as it depends on chronologically consecutive rows of data. The two sequence lengths considered were 3 previous rows of data and 7 previous rows of data. The best accuracy for LSTM was 94.82\% obtained from the ROC+BUF+SYR+ALB, normalized with MinMax, using a sequence length of 3.

\subsubsection{Overall Classification Results}
\begin{figure}[htb]
{
\centering
\includegraphics[width=0.8\textwidth]{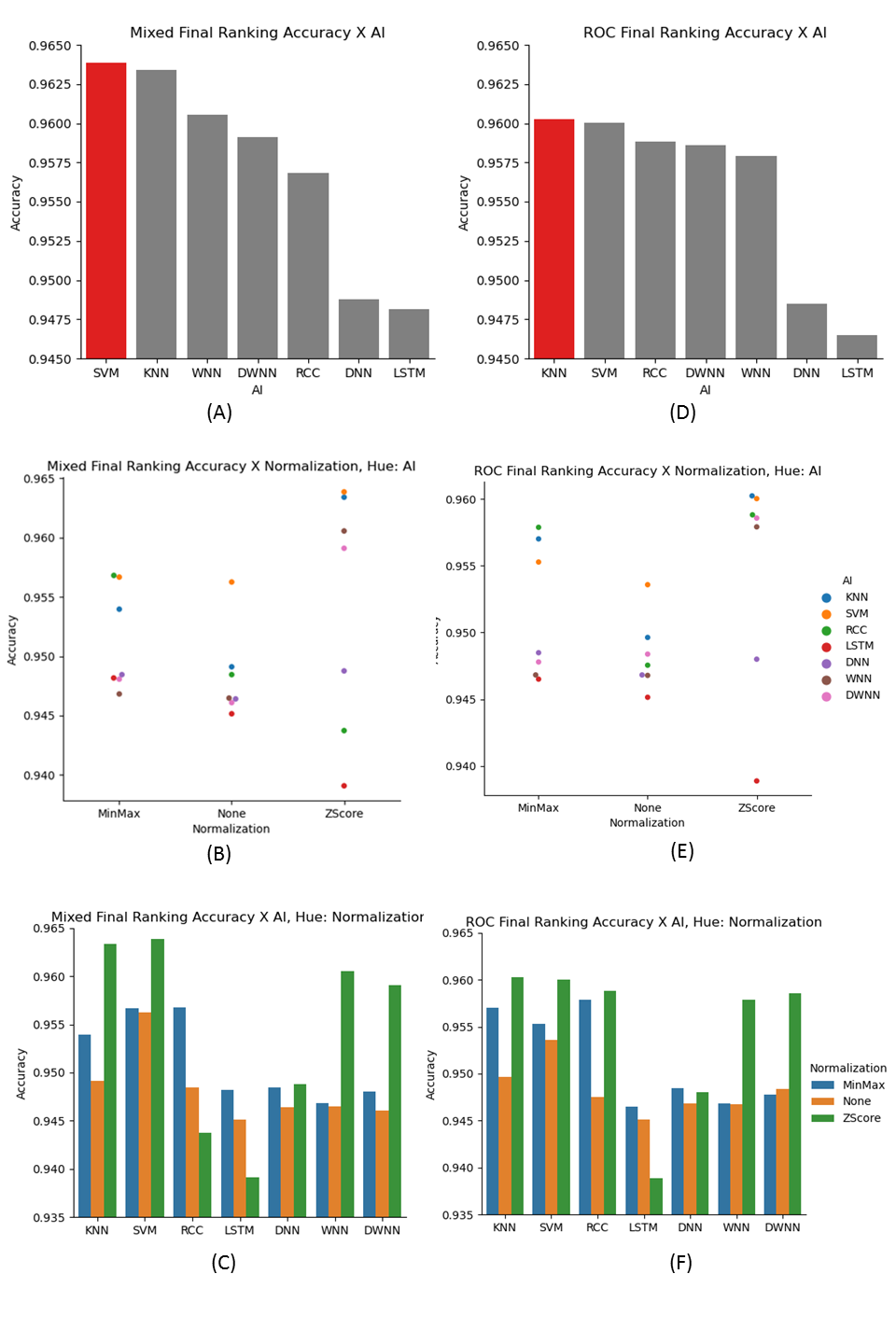}
\caption{(A)-(C) Classification ranking in the Mixed dataset. (D)-(F) Classification ranking in the ROC dataset. (A) and (D) The best performance ranking mixing the normalization. (B) and (E) Accuracy with normalization. (C) and (F) An overall ranking among models and normalizations. }\label{fig:6in1}
}
\end{figure}

Overall, both SVM and KNN classifiers performed the best. \autoref{fig:6in1} (A) and (D) shows all of the best accuracies for each of the classification methods used in the mixed and ROC datasets. The plotting for this ranking was zoomed into the accuracy range of 94.5\%-96.5\% to better show the differences between the methods. A further examination of the of the ranking with the normalization data is shown in \autoref{fig:6in1} (C) and (F). In almost every case except for LSTM, the best normalization method was ZScore. It can also be seen that in many cases the non-normalized version of the data performed measurably worse than its ZScore counterpart. \autoref{fig:6in1} (B) and (E) shows another way to see that same data that shows more clearly the performance in regards to normalization.

\subsection{Regression Models}
Being that accuracy is not measured the same for regressors the metrics recorded were $R^2$, Mean Square Error (MSE), Root Mean Square Error (RMSE), and Pearson Correlation Coefficient (PCC). \par
Some regressors used Numpy’s Random which was creating results that were not reproducible. By setting a Numpy Random Seed prior to training, results became reproducible. A quick and simple investigation was done by iterating through 1-100 for Numpy’s Random Seed. Each iterations seed was used to train and evaluate a simple Linear Regression model. The results of that investigation were that a Numpy Random Seed of 46 resulted in the best $R^2$ value for a Linear Regression Model. Note that this is separate from the random states used in training methods.\par
The data used to train and test the regressors in this section was a subset of the data used for the classifiers such that all rows targets were above 0.01, that target being the next hours p01i value (Precipitation inches). This decision was made after an initial investigation with the full dataset resulted in much poorer performances. This was likely due to the overwhelming ration of 0 precipitation hours compared to non-0 precipitation hours. \par

\subsubsection{KNN Regressor}
Each of the 6 datasets was crossed with the same 3 random states as before to make 18 unique forms of data and then each of those 18 was run through a KNN Regressor 1000 times to find the n\_neighbor parameter that resulted in the best $R^2$ metrics. The dataset that had the best $R^2$, 0.2095615, was the Rochester, Buffalo, Syracuse, and Albany dataset, normalized using ZScore, using None for Random State, and with an n\_neighbors value of 42. This dataset also had the best Pearson Correlation score, 0.46443163. The dataset that had the best MSE and RMSE, 0.005347 and 0.073122, was the Rochester dataset, normalized using MinMax, using Random State of None, and a n\_neighbor value of 51. \autoref{tab:knnr} shows the top 2 best metric performers in KNN regressor. 

\begin{table}[htb]
\caption{Top 2 best metric performers in KNN regressor.}\label{tab:knnr}
\centering
\begin{tabular}{rlllllll}
\hline
Dataset & Normalization & Random & Neighbor & $R^2$ & MSE & RMSE & Pcc\\ \hline
Mixed & ZScore & none & 42 & 0.209561502 & 0.006806884 & 0.082503842 & 0.464431631\\
Mixed  & MinMax & None & 38 & 0.164908114 & 0.00707709 & 0.084125443 & 0.418229929\\
ROC & MinMax & None & 51 & 0.168353653 & 0.005346776 & 0.07312165 & 0.411062473\\
ROC & None & None & 65 & 0.106086597 & 0.006163514 & 0.078508053 & 0.326574449\\
\hline
\end{tabular}
\end{table}

\subsubsection{Linear Regressor}
Each of the 6 datasets were crossed with the same 3 random states as before and run through a linear regression model. The best metrics were obtained from the Rochester dataset, not normalized, and a random state of None. Those values were $R^2$ = 0.178324, Pearson Correlation = 0.422666, MSE = 0.006913, and RMSE = 0.083146. \autoref{tab:linear} shows the top two performing Linear Regression Models and their respective performance metrics.

\begin{table}[htb]
\caption{Top 2 best metric performers in KNN regressor.}\label{tab:linear}
\centering
\begin{tabular}{rllllll}
\hline
Dataset & Normalization & Random  & $R^2$ & MSE & RMSE & Pcc\\ \hline
Mixed & ZScore & none & 0.1464424 & 0.007025658 & 0.083819199 & 0.388019644\\
Mixed  & MinMax & None  & 0.13850212 & 0.008926641 & 0.094480901 & 0.375745596\\
ROC & None & None  & 0.178324011 & 0.006913207 & 0.083145696 & 0.422665921\\
ROC & ZScore & None  & 0.143381358 & 0.009098939 & 0.095388361 & 0.380159271\\
\hline
\end{tabular}
\end{table}

\subsubsection{SVM Regressor}
Each of the 6 datasets was crossed with the same 3 random states as before and crossed again with the 4 different kernel options: linear, poly, rbf, and sigmoid. This resulted in 72 data forms that were run through the support vector regressor. \par
\autoref{tab:svr} shows the best performers in each metric. The best $R^2$, 0.051143, came from the Rochester, Buffalo, Syracuse, and Albany dataset, normalized with z-score, using the rbf kernel, and with a random state of None. However, SVR did not perform well on ROC dataset where the invalid values were calculated in $R^2$. \autoref{tab:svr} shows the top 2 best performers in SVR including the two invalid values in $R^2$.

\begin{table}[htb]
\caption{Top 2 best metric performers in SVR.}\label{tab:svr}
\centering
\begin{tabular}{rlllllll}
\hline
Dataset & Normalization & Kernel & Random & $R^2$ & MSE & RMSE & Pcc\\ \hline
Mixed & ZScore & rbf & none & 0.051143241 & 0.013033322 & 0.114163574 & 0.314971336\\
Mixed  & MinMax & rbf & 42 & 0.036319941 & 0.011336791 & 0.106474368 & 0.380176974\\
ROC & None & sigmoid & 42 & invalid & 0.012855096 & 0.113380316 & 0.168736666\\
ROC & None & sigmoid & 0 & invalid & 0.011151403 & 0.105600204 & 0.171740467\\
\hline
\end{tabular}
\end{table}

\subsubsection{DNN Regressor}
In total, there were 126 DNN Regressors were trained against the cross of the 6 premade datasets, 7 different amounts of hidden layers, and the 3 Random States used from before. Seven different hidden layers were: 2, 3, 4, 5, 10, 20, and 30 layers. Each of the hidden layers had the number of inputs equal to the number of features which for the ROC datasets was 9 and for the ROC+BUF+SYR+ALB there were 36 inputs at each hidden layer. \par


\autoref{tab:dnnr} shows the best two performers in each dataset. The best $R^2$, 0.038945, came from mixed dataset, normalized with MinMax, using 5 hidden layers, and with a random state of None. The best Pearson Correlation metric, 0.267228, came from the mixed dataset, normalized with MinMax, using 3 hidden layers, and with a random state of None. Even though we used 30 layers deep neural network, the best performer of $R^2$ in ROC dataset was obtained from the normalization with MinMax, using 4 hidden layers, and a random state of 42. \par

\begin{table}[htb]
\caption{Top 2 best metric performers in DNNr.}\label{tab:dnnr}
\centering
\begin{tabular}{rlllllll}
\hline
Dataset & Normalization & Layers & Random & $R^2$ & MSE & RMSE & Pcc\\ \hline
Mixed & MinMax & 5 & none & 0.038944969 & 0.010208255 & 0.101035909 & 0.219027928 \\
Mixed  & MinMax & 10 & none & 0.0020580 & 0.00821409 & 0.09063162 & 0.05852291\\
ROC & MinMax & 4 & 42 & 0.036257134 & 0.01133753 & 0.106477838 & 0.24003999\\
ROC & MinMax & 5 & none & 0.035922961 & 0.010334113 & 0.101656837 & 0.191548407\\
\hline
\end{tabular}
\end{table}

\subsubsection{WNN Regressor}

Similarly to the WNN Classifier, there were 18 unique WNN Classifiers by training against the cross of the 6 premade datasets and the 3 Random States used from before. For each of the measured metrics there were different models that performed best. \autoref{tab:wnnr} shows the best performers in each metric. The best $R^2$, 0.132945, came from the Rochester, Buffalo, Syracuse, and Albany dataset, normalized with ZScore, and with a random state of None. The best Pearson Correlation metric also came from the mixed dataset, normalized with MinMax, and with a random state of 42 (not listed on the table). The best MSE/RMSE values were obtained from the mixed dataset normalized with MinMax, and a random state of None (not listed on the table). Those values were MSE = 0.007244 and RMSE = 0.085110 with $R^2$=0.119940624
, and Pcc=0.357592781. In ROC dataset, the best $R^2$ is obtained with a decent MSE and RMSE.

\begin{table}[htb]
\caption{Top 2 best metric performers in WNN regressor.}\label{tab:wnnr}
\centering
\begin{tabular}{rllllll}
\hline
Dataset & Normalization & Random  & $R^2$ & MSE & RMSE & Pcc\\ \hline
Mixed & ZScore & none & 0.132945285 & 0.008984219 & 0.094785122 & 0.368640036\\
Mixed  & MinMax & 0  & 0.121693571 & 0.008829052 & 0.093963034 & 0.354665963\\
ROC & ZScore & 42  & 0.123917918 & 0.010306283 & 0.101519866 & 0.352715523\\
ROC & MinMax & None  & 0.121567081 & 0.009098939 & 0.096595284 & 0.353484533\\
\hline
\end{tabular}
\end{table}

\subsubsection{DWNN Regressor}
Similarly to the DNN Regressor, 7 different hidden layers were: 2, 3, 4, 5, 10, 20, and 30 layers. Table 6 shows the best performers in each metric. The best $R^2$ for the mixed datasets, 0.181974, came from the mixed dataset, normalized with ZScore, using 20 hidden layers, and with a random state of 42. The best $R^2$ for the ROC datasets, 0.143873, were obtained from the Rochester dataset normalized with ZScore, using 20 hidden layers, and a random state of None.


\begin{table}[htb]
\caption{Top 2 best metric performers in DNNr.}\label{tab:dwnnr}
\centering
\begin{tabular}{rlllllll}
\hline
Dataset & Normalization & Layers & Random & $R^2$ & MSE & RMSE & Pcc\\ \hline
Mixed & ZScore & 20 & 42 & 0.181974 & 0.009623 & 0.098098 & 0.426754 \\
Mixed  & ZScore & 30 & 42 & 0.181682 & 0.009627 & 0.098116 & 0.428402 \\
ROC & MinMax & 20 & none & 0.143873 & 0.008235 & 0.090748 & 0.379773 \\
ROC & MinMax & 30 & none & 0.108191 & 0.004529 & 0.067294 & 0.361150 \\
\hline
\end{tabular}
\end{table}

\subsubsection{LSTM Regressor}

There were totally 30 LSTM Regressors by training against the cross of the 6 premade datasets, and 5 different sequence lengths. Random states were not used for LSTM as it depends on chronologically consecutive rows of data. The 5 sequence lengths considered were 3, 5, 7, 9, and 12 previous rows of data.
\autoref{tab:lstm} shows the results of the two best performing LSTM regressors over two datasets. The best $R^2$, MSE, and RMSE came from the same model, using the Rochester dataset, not normalized, and using a sequence length of 5. Those values were $R^2$ = 0.099439, MSE = 0.006424, and RMSE = 0.080149. The best Pearson Correlation, 0.324997, came from the Rochester dataset, normalized using ZScore, and using a sequence length of 3. From the mixed dataset, the regressor with highest $R^2$ obtains the lowest MSE and RMSE and the highest Pcc.  

\begin{table}[htb]
\caption{Top 2 best metric performers in LSTM.}\label{tab:lstm}
\centering
\begin{tabular}{rlllllll}
\hline
Dataset & Normalization & Sequence  & $R^2$ & MSE & RMSE & Pcc\\ \hline
Mixed & MinMax & 12  & 0.093923238 & 0.006489071 & 0.080554771 & 0.310782912 \\
Mixed  & None & 12  & 0.072728045 & 0.006640865 & 0.081491504 & 0.272888933 \\
ROC & None & 5  & 0.099438992 & 0.006423882 & 0.080149126 & 0.318826 \\
ROC & ZScore & 12  & 0.088653463 & 0.006526812 & 0.080788686 & 0.313515598 \\
\hline
\end{tabular}
\end{table}

\subsubsection{LSTM Bi-direction}
The Bidirectional Regressor models used the same settings as the LSTM Regressor models, so 30 unique Bidirectional Regressors were examined. \autoref{tab:lstmbi} shows the results of the two best performing LSTM regressors. The best $R^2$, MSE, RMSE, and Pcc in the Rochester dataset came from the same regressor,  normalized using MinMax, and using a sequence length of 12. Those values were $R^2$ = 0.092928, MSE = 0.006496, and RMSE = 0.080599. The best $R^2$, MSE, RMSE, and Pcc in the mixed dataset also came from the same regressor that was normalized using MinMax, and using a sequence length of 3.

\begin{table}[htb]
\caption{Top 2 best metric performers in LSTM Bi-direction.}\label{tab:lstmbi}
\centering
\begin{tabular}{rlllllll}
\hline
Dataset & Normalization & Sequence  & $R^2$ & MSE & RMSE & Pcc\\ \hline
Mixed & MinMax & 3  & 0.067291762 & 0.006645372 & 0.081519154 & 0.29581004 \\
Mixed  & MinMax & 7  & 0.06109682 & 0.006705727 & 0.081888505 & 0.28511238 \\
ROC & MinMax & 12  & 0.09292772 & 0.006496201 & 0.080599012 & 0.320602646 \\
ROC & None & 12  & 0.085119002 & 0.006552125 & 0.080945195 & 0.311874173 \\
\hline
\end{tabular}
\end{table}
\par

\subsubsection{GRU Regressor}
The GRU Regressor models used the same settings as the LSTM and Bidirectional Regressor models, so 30 unique GRU Regressors. The best metrics in ROC dataset were normalized using ZScore, and a sequence length of 3. Those values were $R2$ = 0.088520, Pearson Correlation = 0.323445, MSE = 0.006494, and RMSE = 0.080586. The best metrics in mixed dataset were normalized using MinMax, and a sequence length of 9. Those values were $R2$ = 0.073344684, Pearson Correlation = 0.293189287, MSE = 0.006625979, and RMSE = 0.081400116. \autoref{tab:gru} shows the top two performing GRU Regressor Models and their respective performance metrics.

\begin{table}[htb]
\caption{Top 2 best metric performers in GRU.}\label{tab:gru}
\centering
\begin{tabular}{rlllllll}
\hline
Dataset & Normalization & Sequence  & $R^2$ & MSE & RMSE & Pcc\\ \hline
Mixed & MinMax & 9  & 0.073344684 & 0.006625979 & 0.081400116 & 0.293189287 \\
Mixed  & MinMax & 5  & 0.067207667 & 0.006653795 & 0.081570797 & 0.29401964 \\
ROC & ZScore & 3  & 0.088519946 & 0.006494126 & 0.080586137 & 0.323445285 \\
ROC & MinMax & 5  & 0.079144128 & 0.00656865 & 0.081047207 & 0.301406376 \\
\hline
\end{tabular}
\end{table}

\subsubsection{  Overall Regression Results}

\autoref{tab:overallregmixed} demonstrates the best performer of each models in the mixed ROC+BUF+SYR+ALB dataset. It shows that KNN gives the best $R^2$ that means that the prediction model best fits the observations at the same time its MSE and RMSE are superior, similar to those metrics in the models of LSTM, GRU, and Bidirectional LSTM, 0.006806884 vs (0.006489071, 0.006625979, and 0.006645372). Also, KNN's Pcc shows the highest value among the models, which means a high correlation coefficient between the prediction model and the dataset. The $R^2$ value of KNN outperforms the rest models. The second place is DWNNr with 20 layers and ZScore normalization. \par  

Recurrent neural network models were expected to have a better performance in sequential data prediction. However, recurrent neural network models such as LSTM, GRU, and bidirectional LSTM have not shown a strong indicator in $R^2$ and Pcc. The experimental results partially match the previous researcher's findings. \cite{liu2020applicability} revealed that the rainfall time series have very short term memory characteristics. It sheds some lights on why LSTM models does not give the best performance in the precipitation forecasting.  \par
   
\begin{table}[htb]
\caption{Overall Performance in ROC+BUF+SYR+ALB dataset}\label{tab:overallregmixed}
\centering
\begin{tabular}{rlllllll}
\hline
Model & Normalization & Random & Parameter  & $R^2$ & MSE & RMSE & Pcc\\ \hline
KNN & ZScore & None  & 42 nodes & 0.209561502 & 0.006806884 & 0.082503842 & 0.464431631 \\
DWNNr	&	ZScore	& 42	& 20	layers & 0.181974116	& 0.009623306	& 0.098098453 & 0.426754455 \\
Linear	&	ZScore	& None	& N/A	& 0.1464424	& 0.007025658	& 0.083819199	& 0.388019644 \\
WNNr	&	ZScore	& None	& N/A	& 0.132945285	& 0.008984219	& 0.094785122	& 0.368640036 \\
LSTM	&	MinMax	& N/A	& 12 units 	& 0.093923238	& 0.006489071	& 0.080554771	& 0.310782912 \\
GRU	 	&	MinMax	& N/A	& 9	units & 0.073344684	& 0.006625979	& 0.081400116	& 0.293189287 \\
Bidirect &	MinMax	& N/A	& 3	units & 0.067291762	& 0.006645372	& 0.081519154	& 0.29581004 \\
DNNr  &	MinMax	& None	& 5	layers& 0.038944969	& 0.010208255	& 0.101035909	& 0.219027928 \\
SVR	& ZScore	& 42	& rbf kernel	& 0.036319941	& 0.011336791	& 0.106474368	& 0.380176974 \\
\hline
\end{tabular}
\end{table}

In the ROC dataset, KNN is ranked the second place with a little lower value in $R^2$ than that in the linear regression model, showing the performance 0.178324011 vs. 0.168353653. However, KNN is much better in MSE and RMSE than those of the first place. Comparing with the third place DWNNr, all metrics of KNN are superior. The ROC dataset is not so big as the mixed dataset that KNN may be a little inferior in $R^2$ and Pcc than the first place but in other metrics KNN exhibited its strong prediction ability in rainfall forecasting. \par  

\begin{table}[htb]
\caption{Overall Performance in ROC dataset}\label{tab:overallregROC}
\centering
\begin{tabular}{rlllllll}
\hline
Model & Normalization & Random & Parameter  & $R^2$ & MSE & RMSE & Pcc\\ \hline
Linear	&	None	& None	& N/A	& 0.178324011	& 0.006913207	& 0.083145696	& 0.422665921 \\
KNN	& MinMax & None & 51 nodes & 0.168353653 & 0.005346776 & 0.07312165 & 0.411062473 \\
DWNNr & ZScore & None & 20 layers & 0.143873162 & 0.008235172 & 0.090747846 & 0.379772590 \\
WNNr & ZScore & 42 & N/A & 0.123917918 & 0.010306283 & 0.101519866 & 0.352715523 \\
LSTM & None	 & N/A	 & 5 units & 0.099438992 & 0.006423882 & 0.080149126 & 0.318826 \\
Bidirect & MinMax & N/A & 12 units & 0.09292772 & 0.006496201 & 0.080599012 & 0.320602646 \\
GRU & ZScore & N/A & 3 units & 0.088519946 & 0.006494126 & 0.080586137 & 0.323445285 \\
DNNr & MinMax & 42 & 4 layers & 0.036257134 & 0.01133753 & 0.106477838 & 0.24003999 \\
SVR & None & 42 & sigmoid & invalid & 0.012855096 & 0.113380316 & 0.168736666 \\
\hline
\end{tabular}
\end{table}

\section{Conclusion and Discussion}
Rainfall forecasting plays an important role in our daily lives, especially agriculture and related activities. We have designed an integrated tool of machine learning modeling for regional rainfall forecasting. It applied the state-of-the-art machine learning algorithms including Deep Neural Network (DNN), Reservoir Computing (RC), Wide Neural Network (WNN), Deep and Wide Neural Network (DWNN), Long Short Term Memory (LSTM), Support Vector Machine (SVM), and K-Nearest Neighbor (KNN). Not only the classification models but also the regression models were implemented. In the classification models, we used the classification models to forecast if the precipitation will occur in next hour. In the regression models, we used the regression models to predict the regression value of the rainfall. Also, we adopted two normalization methods MinMax and ZScore and compared their performance with those of non-normalized models. The results shows that KNN model with ZScore normalization can outperform over other models according to the metrics of $R^2$, MSE, RMSE, and Pcc. \par 

Time series prediction models such as LSTM, GRU, and LSTM-bidirection have not shown satisfactory performance in rainfall forecasting. On the contrary, the linear based models such as DWNN, linear models, and WNN have much better performance than complicated models such as DNN, SVR, and LSTM. The reason behind the ranking is that the weather data unlike other data contains many uncertain outliers, errors, and missings that may fuel lots of distractions to the precise models such as neural networks, and time series models. Therefore, rainfall forecasting requires more robust models to analyze those weather datasets. Fortunately, through the experimental results, we can find that the KNN model, a model that has been underestimated, is robust enough to be qualified for this type of job. In the future, we will focus on refining the rainfall forecasting models, especially those methods that can detect the anomaly in order to resist their impact on weather datasets.  \par

\bibliographystyle{scsproc}
\bibliography{scs21paper-nyu}

\end{document}